\pdfoutput=1

\documentclass[11pt]{article}
\usepackage[]{EMNLP2022}

\usepackage[T1]{fontenc}

\usepackage{algorithm}
\usepackage{algorithmic}
\usepackage{times}
\usepackage{latexsym}
\usepackage[utf8]{inputenc}
\usepackage{microtype}
\usepackage{xspace}
\usepackage{caption}
\usepackage{subcaption}
\usepackage{amsmath}
\usepackage{url}
\usepackage{amssymb}
\usepackage{bbding}
\usepackage{amsfonts}
\usepackage{graphicx}
\usepackage{tabularx}
\usepackage{multirow, multicol}
\usepackage{arydshln}
\usepackage{mathtools,nccmath}
\usepackage{enumitem}
\usepackage{todonotes}
\usepackage{cleveref}
\usepackage{booktabs}
\usepackage{color}
\usepackage{hyperref}

\newcommand{\tf}[1]{\textbf{#1}}
\newcommand{\ti}[1]{\textit{#1}}
\newcommand{\ul}[1]{\ti{\underline{#1}}}
\newcommand{\tif}[1]{\textit{\textbf{#1}}}

\newcommand{\ourmodel}{\textsc{ReGrouP}\xspace}
\newcommand{\knowledge}{formulaic knowledge\xspace}
\newcommand{\kit}{knowledge-intensive text-to-SQL\xspace}

\newcommand{\fsp}{\textsc{KnowSQL}\xspace}

\title{Towards Knowledge-Intensive Text-to-SQL Semantic Parsing with Formulaic Knowledge}

\author{Longxu Dou$^{1}$\thanks{~ Contribution during the internship at Microsoft Research Asia.
}, Yan Gao$^{2}$, Xuqi Liu$^{1}$, Mingyang Pan$^{1}$, Dingzirui Wang$^{1}$, \\ \textbf{Wanxiang Che$^{1}$, Min-Yen Kan$^{3}$, Dechen Zhan$^{1}$, Jian-Guang Lou$^{2}$} \\ 
$^{1}$    Harbin Institute of Technology\
$^{2}$    Microsoft Research Asia\\
$^{3}$    National University of Singapore\\
{ \{lxdou, xqliu, mypan, dzrwang, car\}@ir.hit.edu.cn},  {\ dechen@hit.edu.cn},    \\
{ \{yan.gao, jlou\}@microsoft.com},
{ kanmy@comp.nus.edu.sg}\\
}

\begin{document}
\maketitle
\begin{abstract}
    In this paper,
    we study the problem of \tif{\kit}, in which domain knowledge is necessary to parse expert questions into SQL queries over domain-specific tables. 
    We formalize this scenario by building a new Chinese benchmark \fsp consisting of domain-specific questions covering various domains.
    We then address this problem by presenting \tif{\knowledge}, rather than by annotating additional data examples.
    More concretely, 
    we construct a formulaic knowledge bank as a domain knowledge base
    %we define three types of \knowledge: {\it calculation}, {\it union} and {\it condition},
    and propose a framework (\ourmodel) to leverage this formulaic knowledge during parsing.
    Experiments using \ourmodel demonstrate a significant $28.2$\% improvement overall on \fsp.

\end{abstract}

\section{Introduction}
Text-to-SQL translates user queries into executable SQL, greatly facilitating interactions between users and relational databases.
Along with the release of large-scale benchmarks~\cite{zhongSeq2SQL2017,yu-etal-2018-spider,yu-etal-2019-cosql,yu-etal-2019-sparc} and developments in model design~\cite{wang-etal-2020-rat,cao-etal-2021-lgesql}, text-to-SQL works are now achieving promising results in both research and practical applications~\cite{zeng-etal-2020-photon}.

However, in the professional application of text-to-SQL, such as in the data analysis of financial reports, models require external knowledge to map the expert query with the domain-specific database. 
Take the financial query for example: {\it What's the EBIT\footnote{EBIT is Earnings Before Interest and Tax, and is calculated as {\it Revenue -- Cost of Goods Sold -- Operating Expenses.}} of Walmart?}, where the underlying database has component columns that can be used to calculate the EBIT.  
We treat this problem as \tif{\kit}, where domain knowledge is highly necessary to parse expert questions over domain-specific tables.
This problem prevents text-to-SQL techniques from being fielded in novel, professional applications to assist the experts in processing data.

Traditional approaches would address this problem by annotating specific question/SQL pairs on a target domain~\cite{wang-etal-2015-Overnight, herzig-berant-2019-detect}. Then such mappings are induced during the training process. This approach does work but has the drawback that any induced information is both {\it fragile} and {\it expertise-heavy}: such knowledge does not port across domains and requires expert knowledge to craft.

\begin{figure}[tb]
    \centering
    \includegraphics[width=1\linewidth]{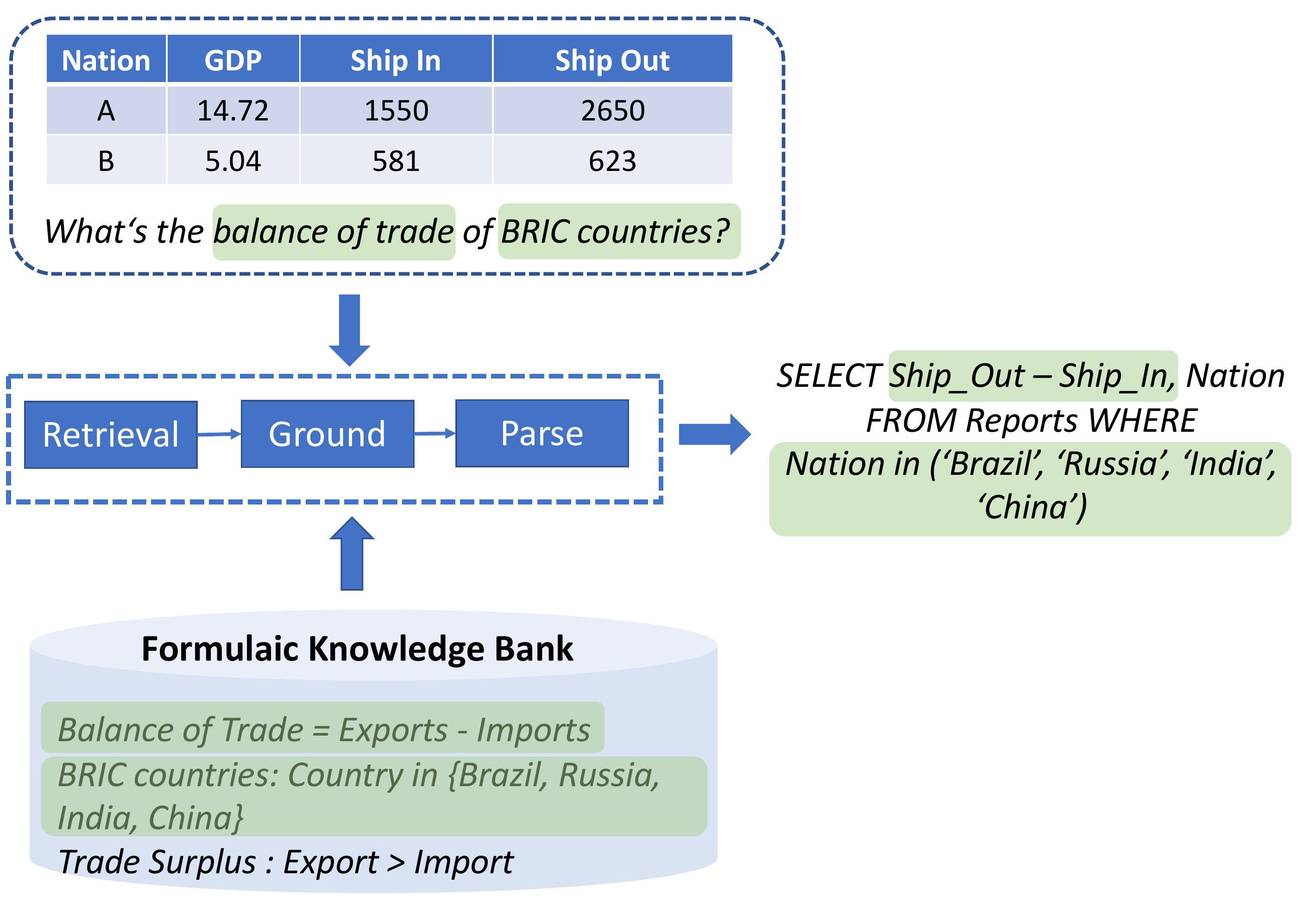}
    \caption{
        Harnessing \ourmodel with \knowledge for \kit with three steps: 
        (1) \tf{Re}trieval the formulaic knowledge; 
        (2) \tf{Grou}nd the concept of formulaic knowledge; 
        (3) \tf{P}arse the question.
% 			The \knowledge is provided by an external formulaic knowledge bank.
    }
    \label{fig:intro}
\end{figure}	

We propose to solve this problem by modeling how a non-expert person might tackle this problem.  When meeting unseen examples (as in the EBIT case above), they may first search for the related mathematical formulas from public resources, then ground the concepts referenced in the formulas with schema elements presented in their particular databases.  This process leverages common, encoded \tif{\knowledge} that are already described in publicly-available resources such as tutorials,   textbooks, encyclopedias, and references. 

Inspired by this, we propose to address the \kit through \tif{\knowledge} which provides the evidence of mapping from domain-specific phrases presented in questions to actual SQL operations over schema elements. 
More concretely, we define a taxonomy of three types of \knowledge: {\it calculation}, {\it union}, and {\it condition}, each corresponding to a particular snippet of SQL.
Then we propose \ourmodel , a text-to-SQL framework (Fig.~\ref{fig:intro}), 
consisting of three stages: 
(1) {\bf Re}trieve the formulaic knowledge from formulaic knowledge bank as an external knowledge source; 
(2) {\bf Grou}nd the concept of formulaic knowledge to the schema elements;
(3) {\bf P}arse the results with the question, schema, and grounded formulaic knowledge.
The external formulaic knowledge bank imbues \ourmodel with \knowledge, making it \ti{knowledgeable}.	
\ourmodel is also \ti{extensible} because updating the formulaic knowledge bank does not require retraining any modules.

Moreover, we construct a Chinese benchmark \fsp, to examine the effectiveness of \ourmodel framework. 
It advances the existing \kit beyond the previous work~\cite{wang-etal-2020-dusql,zhao-etal-2022-bridging} by considering more SQL operations and challenging domains.
Experimental results demonstrate the \ourmodel with \knowledge would improve the performance by $23.4$\% overall. 
Furthermore, we classify error cases into three classes, which are resolvable by advancing the corresponding module of \ourmodel.
Finally, we discuss the potential future work such as expanding the scope of knowledge and advancing \ourmodel model design.

Our contributions are summarised as follows:
\begin{itemize}
    \item To the best of our knowledge, we are the first to explore \kit and propose a challenging Chinese benchmark \fsp, which requires domain-specific knowledge.
    \item We propose a novel framework \ourmodel to address \kit by retrieving and grounding \knowledge, which is knowledge-extensible. 
    \item Experimental results demonstrate the effectiveness of \ourmodel with \knowledge which achieves $28.2$\% overall improvement on \fsp. 
\end{itemize}

\section{Knowledge-Intensive Text-to-SQL}

\subsection{Problem Analysis}
After studying the real cases in professional data analysis, we roughly categorize the required knowledge for \kit into three classes :
(1) \ti{linguistic knowledge} enables the model to adapt to linguistic diversity;
(2) \ti{domain knowledge} allows the model to perceive domain-specific sayings and concepts;
(3) \ti{mathematical knowledge} yields the specific SQL operations (\ti{e.g.}, \ti{Density} phrase to \ti{division} operation).
These three sets of knowledge jointly provide the evidence of \ti{mapping from domain-specific phrases of questions to actual SQL operations over schema elements}.

However, most text-to-SQL researches focus on general scenario~\cite{yu-etal-2018-spider, zhongSeq2SQL2017}, where linguistic knowledge is mainly required.
Recently, \citet{wang-etal-2020-dusql} and \citet{zhao-etal-2022-bridging} promote text-to-SQL to more challenging scenarios via involving the calculation questions.
In this paper, we further explore the \kit by considering more operations (\ti{e.g.}, calculation, union, and condition) with more challenging domains which require all these three classes of knowledge.

\subsection{Challenge}
Despite that pre-trained language models contain linguistic knowledge,
they lack domain knowledge and mathematical knowledge. 
Therefore, the model would meet two problems:
(1) \tif{don't know which operations to use}: if an operation (\ti{e.g.}, \ul{density = total number / space}) has never occurred in training data, the model rarely employ this unseen operation during the inference; 
(2) \tif{don't know how to adapt operations}: 
the model would fail to generalize the operation across domains.
For instance, the model cannot generalize the calculation of \ul{Population Density (number of people / land area)} to \ul{Car Density (number of cars / parking lot area)}.

Accordingly, we consider that the vanilla pre-trained language model is (1) \tif{narrow} since it only supports the limited operation and
(2) \tif{inefficient} since it can't generalize the operation across domains.
However, it's time-consuming and expertise-heavy to directly increase the amount of annotated data examples.
In contrast, we address this challenge from the view of \knowledge in Sec~\ref{sec:formulaic_knowledge}, which is more knowledge-extensible.

\subsection{\fsp Benchmark}

%statistic_of_benchmark
\begin{table}[htp]
    \centering
    \small
    \begin{tabular}{lcccc}
        \toprule
        {} & \bf \#DB & \bf \#Question & \bf \#Formulaic  \\
        \midrule
        {\bf Train} & {$160$} & {$23,157$}  & {$328$} \\
        {\bf Dev} & {$40$} & {$2,731$}  & {$122$} \\
        \hdashline
        {\bf Finance} & {$217$} & {$1,392$}  & {$219$}\\
        {\bf Estate} & {$35$} & {$749$}  & {$79$}\\
        {\bf Transportation} & {$36$} & {$439$}  & {$82$}\\
        \bottomrule
    \end{tabular}
    \caption{The dataset statistic of \fsp. 
    } 
    \label{tab:statistic_of_knowsql}
\end{table}

To uncover the knowledge-intensive text-to-SQL problem and advance the research, we construct a challenging Chinese text-to-SQL benchmark named \fsp. 
Roughly, it consists of two parts: training/dev sets built on the existing DuSQL~\cite{wang-etal-2020-dusql} dataset and a newly constructed test set on three professional domains with discovered knowledge in DuSQL.

\subsubsection{Building Training/Dev Set on DuSQL}
We build the training/dev set of \fsp based on the existing DuSQL, a Chinese multi-table text-to-SQL benchmark. 
We categorize its 200 databases into 16 domains like sports, energy, health care, foods, \ti{etc}.
Given the high quality of DuSQL schema and broad domain coverage, it's a satisfactory start-point to build a challenging \kit benchmark.
However, the domain-specific question is not well included in DuSQL, where most of the questions could be answered easily without relying on external knowledge and only considers one SQL operation (\ti{i.e.}, calculation).
Given that, we extend the original DuSQL by adding more domain-specific questions and involving more operations in both the train set and the dev set.
Eventually, \fsp expands the size of DuSQL train set from 22,521 to 23,157 and the dev set from 2,482 to 2,731.

\subsubsection{Building Test Set from Scratch}
To simulate the professional data analysis scenario, we create a challenging test set covering three domains (finance, estate, and transportation).
These three domains have high data analysis requirements in real life.
Different from the train/dev sets, we construct the test set from the scratch by: (1) collecting the domain-specific tables, and (2) annotating the domain-specific questions and corresponding SQL queries.

\paragraph{Table Collection.} 
For collecting table schema, we collect the tables from the following source: (1) the public annual reports of the company (2) the industry reports (3) academic papers (4) the statistical reports released by the government.
To ensure the table quality, we conduct several pre-processing procedures.
Firstly, we convert matrix tables (present in annual reports) into relational tables to make the question SQL-answerable. Next, to ensure the table data quality, we conduct data cleaning (\ti{e.g.}, filtering out the irrelevant columns to simplify the table structure, and normalizing the headers to reduce the noise). 
Finally, to avoid data privacy issues, we conduct value anonymization (\ti{e.g.}, removing direct identifiers and anonymizing geo-related data). 

\paragraph{Question Annotation.} 
It’s challenging for annotators to propose the domain-specific questions without background knowledge~\footnote{See Sec.~\ref{sec:ethical_considerations} for annotator payment and profile.}. 
Thus, we train the annotators first about the domain-specific knowledge via 
(1) collecting the jargon (\ti{i.e.}, abbreviation, terminology) from the domain-specific open resources, which are widely adopted by domain experts (e.g., EBIT for finance) but unusual for a layperson; 
(2) to mimic the domain expert by asking questions using the jargon with the above materials.

After that, the annotators would annotate the questions and SQL with the following criteria:
(1) be faithful to the given table (\ti{i.e.}, don’t exceed the scope of table columns and table content); 
(2) not be directly answerable by the single element of the table but could be answered by the operation over existing columns; 
(3) limited to first-order operation (\ti{i.e.}, excludes multi-hop questions like ‘What is the gross profit?’, where the table only contains ‘Sales’, ‘Average Price’ and ‘Cost of Goods Sold’ so that model needs to compute the ‘revenue’ first). 

\subsubsection{Dataset Quality and Data Statistic}\label{sec:dataset_statistic}
To guarantee the data quality, we conduct a multi-rounds check. Finally, the inter-agreement of annotators reaches 94.7\% \footnote{The inter-annotator agreement is calculated as the percentage of overlapping votes about whether it's a correct and domain-specific question.}.
During each round, we ask each annotator to review others’ annotations based on the criteria (stated above), then ask them to further improve annotations that do not meet the criteria. 
    As shown in Tab.\ref{tab:statistic_of_knowsql}, the test set contains \tif{288} databases and \tif{2,580} questions.
Notably, all these challenging data examples in the test set could be covered by \tif{380} formulaic knowledge, which will be discussed in Sec.~\ref{sec:formulaic_knowledge}.

%ebit
\begin{figure}[tb]
    \centering
    \includegraphics[width=1\linewidth]{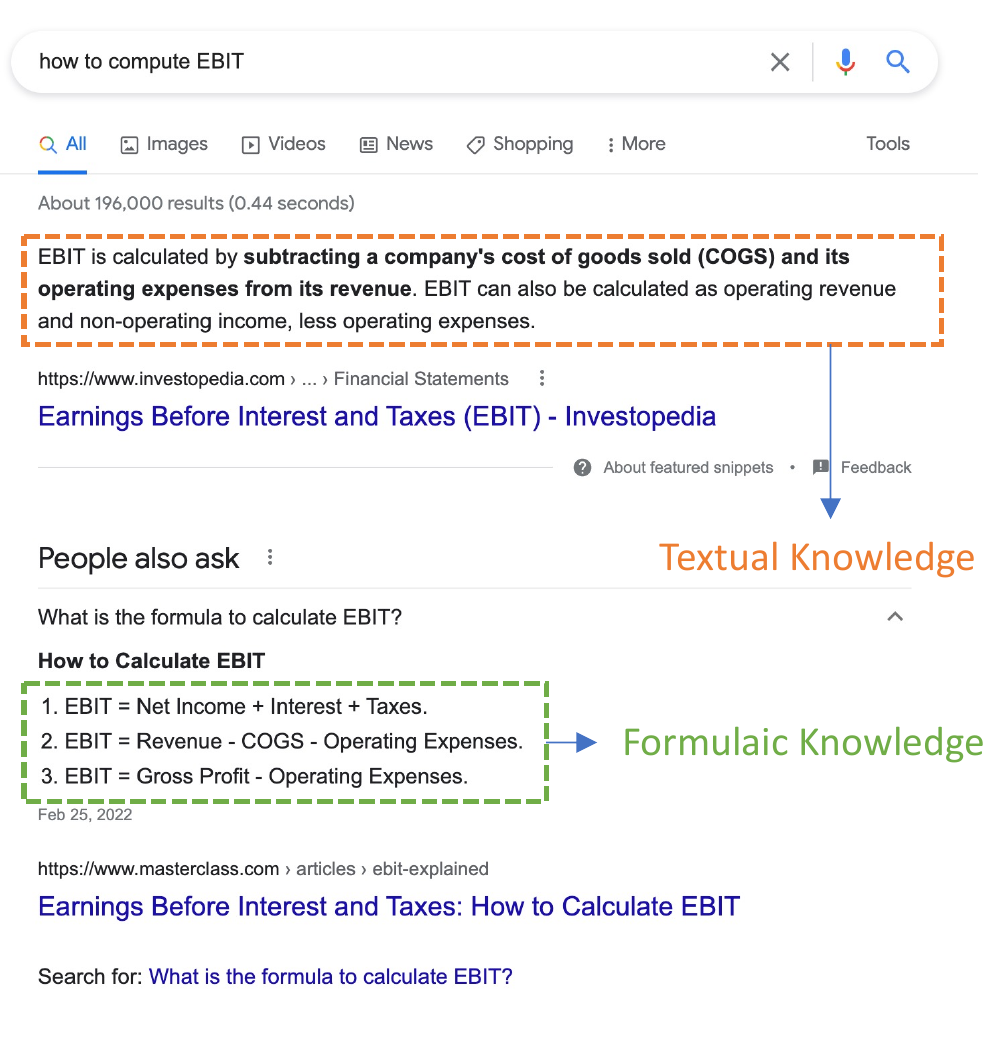}
    \caption{
        Two types of knowledge in expressing \ti{the calculation of EBIT}: \tif{textual knowledge} and \tif{formulaic knowledge}.
    }
    \label{fig:ebit}
\end{figure}

\section{Approach: Formulaic Knowledge}\label{sec:formulaic_knowledge}

\subsection{Motivation}\label{sec:motivation}
When meeting unseen examples, the human may first search the related mathematical knowledge or domain knowledge from textbooks or encyclopedias.
As shown in Fig.~\ref{fig:ebit}, the information of calculation of EBIT is returned in both textual and formulaic format.
Intuitively, the formulaic format is preferred because it's 
(1) \tif{more concise and precise}: for instance, \ul{a adds b times c} is more ambiguous than \ul{a+b*c} or \ul{(a+b)*c};
(2) \tif{easy to obtain}: most description of calculation is stored in a formulaic format in the textbook, tutorials, and academic paper;
(3) \tif{SQL parser friendly}: the formulaic format is closed to the snippet of SQL then easily for the parser to generate\footnote{In Sec.~\ref{sec:discussion}, experimental results prove that formulaic format receives better performance than textual format.}.

%knowledge type
\begin{figure*}[htb]
    \centering
    \includegraphics[width=1\linewidth]{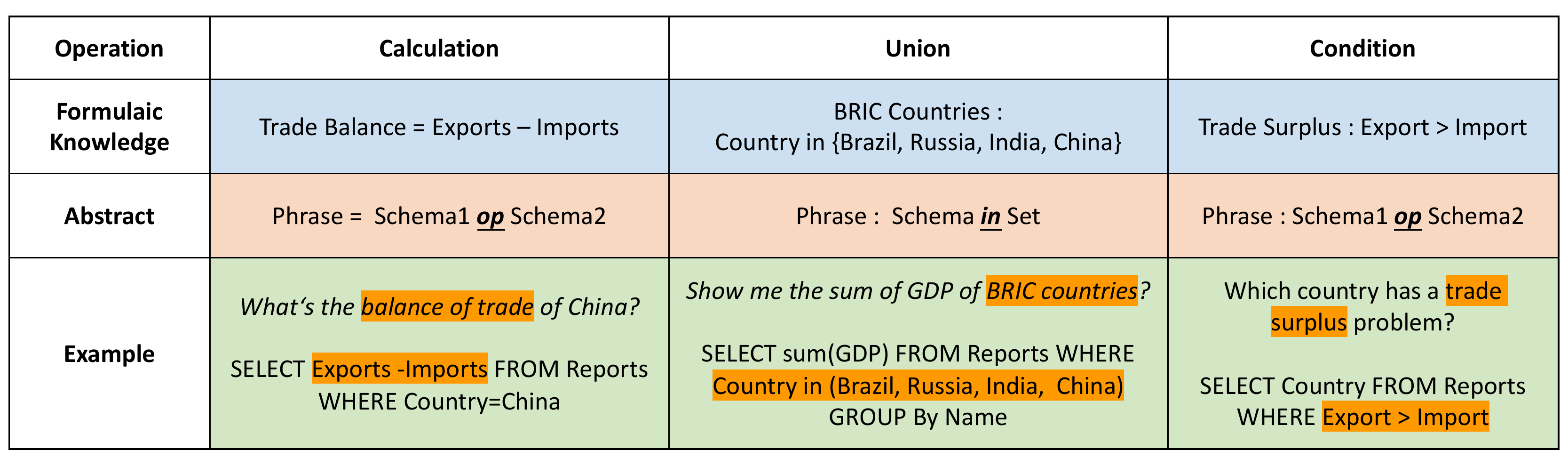}
    \caption{
        We consider three types of formulaic knowledge to address \kit.
    }
    \label{fig:knowledge_type}
\end{figure*}

%statistic_of_bank
\begin{table*}[htp]
    \centering
    \small
    \begin{tabular}{lccccc}
        \toprule
        {}  & \bf \#Formulaic & \bf \#Calculation &\bf \#Union & \bf\#Condition   \\
        \midrule
        {\bf Formulaic Knowledge Bank} & {$1,954$}  & {$1,102$}  &  {$346$} & {$506$}  \\
        {\bf \fsp involved} & {$891$}  & {$656$} & {$52$} & {$183$}   \\
        \bottomrule
    \end{tabular}
    \caption{The dataset statistic of \knowledge bank and its overlap with \fsp.
    } 
    \label{tab:statistic_of_bank}
\end{table*}

\subsection{Formulaic Knowledge for Text-to-SQL}
Following this idea, we focus on three categories of operations (Fig.~\ref{fig:knowledge_type}): calculation, union, and condition.
Besides the popular \ti{calculation knowledge}, we also consider the taxonomy information as \ti{union knowledge} and the judgment standard as \ti{condition knowledge}.
The design insight here is that the left part is the name of the knowledge item, and the right part expresses its semantic meaning represented by operations over concepts.
Note that all operations are consistent with SQL grammar,  making it closer to SQL query.
Besides the entity, the left part of formulaic knowledge might also be the SQL function (\ti{e.g.}, NOW()) or constant (\ti{e.g.}, threshold of Real Estate Bubble).

\subsection{Formulaic Knowledge Bank}

We further build a formulaic knowledge bank with 1,954 formulaic knowledge items, which supports 19 domains involved in \fsp. 
Importantly, the bank covers all these examples of \fsp as shown in Tab.~\ref{tab:statistic_of_bank}.
Note that this bank is a domain-related resource, not one tied to the specific database.
Thus, this bank is more general and could be utilized in other applications natural language applications (\ti{e.g.}, question answering)
\footnote{See fine-grained statistic of bank in Appendix~\ref{app:detail_data}.}.

\paragraph{Criteria}
The design of the formulaic knowledge follows three criteria: (1) Only the first-order (flat) formulaic knowledge is considered (\ti{i.e.}, the concept in the formulaic item should be align-able to the schema elements rather than another formulaic item) ;
(2) The stored formulaic knowledge should be both faithful (\ti{i.e.}, acknowledged by the expert) and standardized (\ti{i.e.}, shared at the domain level);
(3) The formulaic knowledge should be domain-level (\ti{i.e.}, not tied to the specific schema elements).

\paragraph{Collection} 
We collect the \knowledge from the following public resource:
(1) Baidu Wenku, the platform where the domain experts usually share the domain knowledge of various domain\footnote{\href{https://wenku.baidu.com/}{https://wenku.baidu.com/}}; 
(2) CNKI, China's largest academic website\footnote{\href{https://oversea.cnki.net/index/}{https://oversea.cnki.net/index/}};
(3) the data analysis websites of a specific domain, like ESPN for sports and Yahoo for finance. 
We also collect some knowledge from the English resource and let annotators translate this domain knowledge into Chinese.

\begin{figure*}[htb]
    \centering
    \includegraphics[width=1\linewidth]{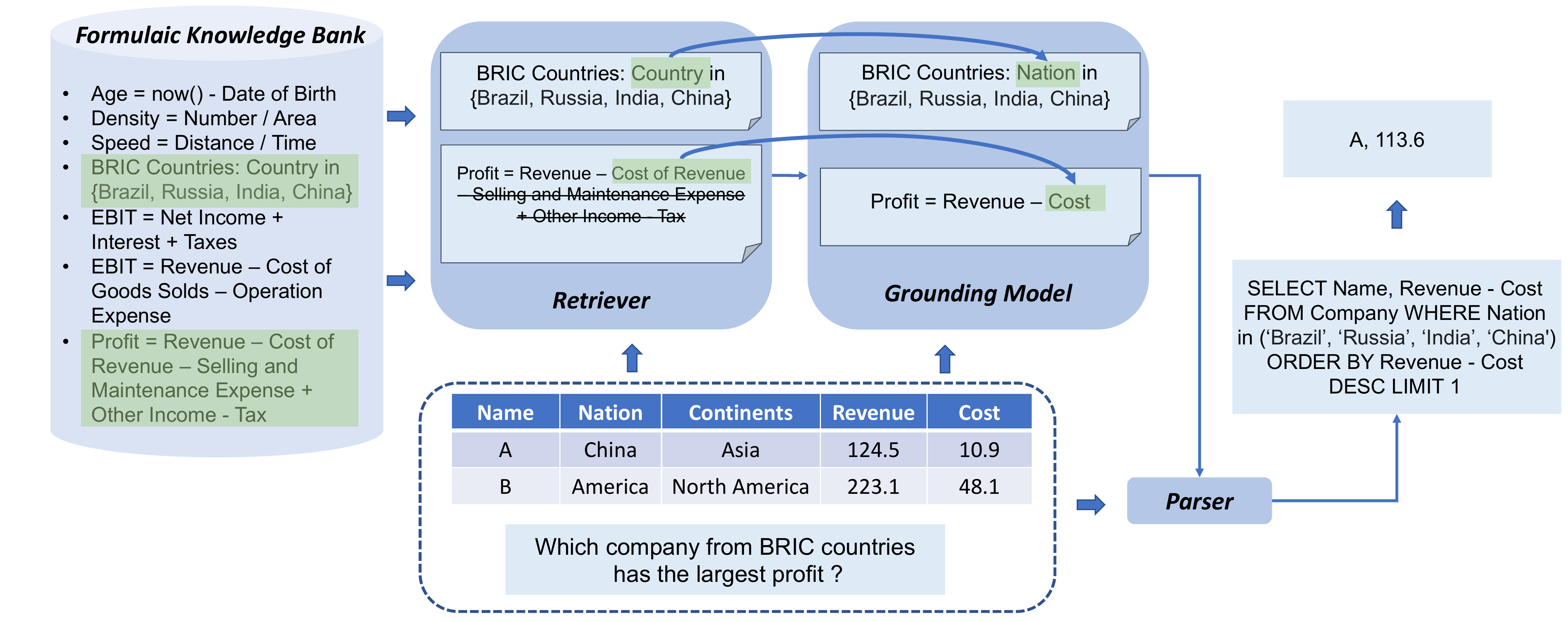}
    \caption{
        Pipeline of \ourmodel:
        (1) \tf{Re}trieve the formulaic knowledge item from the bank;
        (2) \tf{Grou}nd the concept of formulaic knowledge into schema elements;
        (3) \tf{P}arse the question with grounded formulaic knowledge into SQL.
    }
    \label{fig:framework}
\end{figure*}

\paragraph{Abstraction} 
To make the formulaic knowledge more generic, we propose to accumulate the \knowledge at the domain level instead of database-specific. 
Specifically, we abstract the concept of \knowledge before storing them in the knowledge bank, which indicates the operation over concept rather than specific schema.
For example, we would extract the formulaic knowledge from
`People Density in China 2020 = total number of Chinese in 2020 / Chinese Land Area' to `People Density = total number of People / Area'.
Consequently, only \tif{ONE} \knowledge is required to address \tif{MANY} schema elements to calculate the density of animals/cars/shops.

\paragraph{Mapping within \fsp}
We further examine the overlap between formulaic knowledge bank and \fsp benchmark.
As stated in Sec~\ref{sec:dataset_statistic}, all questions from \fsp are covered by formulaic knowledge banks. 
Specifically, there are 1,954 knowledge items in the bank, and 891 items are used for answering the \fsp questions as shown in Table~\ref{tab:statistic_of_bank}. 
Especially, there are extra 1,063 knowledge items beyond \fsp which could support future work in applying formulaic knowledge.

\section{\ourmodel Framework}
To address the \kit problem, we propose a novel framework named \ourmodel, consisting of three stages:
(1) {\bf Re}trieve the formulaic knowledge from the formulaic knowledge bank as an external knowledge source; 
(2) {\bf Grou}nd the concept of formulaic knowledge to the schema elements (\ti{e.g.}, \ti{Exports} to \ti{Ship\_Out});
(3) {\bf P}arse the results with the question, schema, and grounded formulaic knowledge.
As shown in Fig.~\ref{fig:framework}, \ourmodel consists of three models: retriever, grounding model, and parser.
We will give a brief introduction of each model in the following\footnote{More details of the model implementation could be found in Appendix~\ref{app:detail_model}.}. 

\subsection{Retriever Model}
The goal of the retriever is to \ti{extract the relevant formulaic knowledge items from the formulaic knowledge bank} (Fig.~\ref{fig:framework}).
The challenge is the fine-grained modeling of the formulaic knowledge to disambiguate the ones with the same intent but differing in operation over concepts, such as calculating EBIT in different ways. 
We directly utilize the off-the-shelf Dense Passage Retriever (DPR)~\cite{karpukhin-etal-2020-dense} which was originally designed for open-domain QA.
It employs a bi-encoder architecture to learn the dense representation of sentences and passages, then it computes the dot-product between the representations as the similarity score. 

To adapt the DPR in the formulaic knowledge retrieval task, we treat the formula knowledge bank as the passage candidate and concatenate the question with flattened schema (separated by special token `|') to enrich the semantics of the question. 
Then we follow the standard DPR training procedure to optimize the bi-encoder.
Specifically, during the training process, we derive the positive knowledge items from \fsp annotation and sample five negative examples from the formulaic knowledge bank.
During the inference process, we first cache the embedding of formulaic knowledge items, then leverage the FAISS algorithm~\cite{faiss} to rank each formulaic knowledge item.

\subsection{Grounding Model}
Given the retrieved knowledge items, the goal of the grounding model is to \ti{edit the formulaic knowledge items w.r.t specific schema} through 
(1) removing the irrelevant concept and
(2) instantiating the concept with the schema elements.
The main challenge is the expensive annotations of grounding (\ti{i.e. supervision}). 
Therefore, the weakly supervised grounding approaches would be more suitable.
Specifically, we leverage the Erasing-then-Awakening (ETA) model proposed by~\citep{liu-etal-2021-awakening}, which was originally designed for grounding the entity from the knowledge base to the entity mentioned in the question.
The output of ETA is a confidence matrix, indicating the possible grounding relations between entity mentions and entities.

To adapt the ETA in the formulaic knowledge grounding task, we treat each knowledge item as the `question' and attempt to figure out which specific schema elements are grounded in the knowledge item.
Specifically, it's determined by a hyper-parameter \ti{$H$} to indicate the threshold of confidence (whether it's grounded and which one it's grounded).
As shown in Fig.~\ref{fig:framework}, we filter the concept (cross outed parts) under the confidence threshold \ti{$H$} and replace the concept with aligned elements (green parts).

%overall
\begin{table*}[htb]
    \small
    \centering
    \begin{tabular}{lccccc}
        \toprule
        \tf{ \centering Model} &	\multicolumn{1}{c}{\textbf{Dev}} & \multicolumn{1}{c}{\textbf{Finance}}  & \multicolumn{1}{c}{\textbf{Estate}} & \multicolumn{1}{c}{\textbf{Transportation}} &
        \tf{Average}\\ 
        
        \midrule
        
        Vanilla & $69.3$ & $8.7$ & $5.7$ & $6.9$ & $22.7$ \\
        \ourmodel (w/o Grounding) & $71.7$ & $38.1$ & $25.1$ & $32.7$ & $41.9$\\

        \ourmodel & $74.6$ & $43.7$ & $46.1$ & $39.1$ & $50.9$\\

        \hdashline
        \ourmodel (Oracle) & $78.4$ & $71.4$ & $84.8$ & $64.7$ & $74.8$\\
        
        \bottomrule
    \end{tabular}
    \caption{Overall results on different \fsp splits. Oracle refers to the use of the oracle formulaic knowledge. The evaluation metric is SQL exact set match. Average indicates the micro-average score of the first four columns. }
    \label{tab:main_reults}
\end{table*}

%retrieve_reults
\begin{table}[tb]
    \small
    \centering
    \begin{tabular}{ccccc}
        \toprule
        \tf{Data}& \tf{Model} &	\multicolumn{1}{c}{\textbf{R@1}} & \multicolumn{1}{c}{\textbf{R@3}}  & \multicolumn{1}{c}{\textbf{R@10}} \\ 
        \midrule
        \multirow{2}{*}{{\parbox{1.5cm}{\centering Dev}}}
        &BM-25 & $67.9$ & $89.1$ & $96.5$\\
        &\ourmodel & $73.0$ & $89.8$ & $96.5$ \\
        
        \midrule
        
        \multirow{2}{*}{{\parbox{1.5cm}{\centering Finance}}}
        &BM-25 & $39.4$ & $66.5$ & $85.9$\\
        &\ourmodel & $46.0$ & $68.1$ & $86.1$ \\	
        
        \bottomrule
    \end{tabular}
    \caption{Results of \ourmodel retriever compared with BM-25 on \fsp dev and finance splits. The evaluation metric is the Recall.}
    \label{tab:retrieve_reults}
\end{table}

%ground_reults
\begin{table}[tb]
    \small
    \centering
    \begin{tabular}{ccccc}
        \toprule
        \tf{Data}& \tf{Model} &	\multicolumn{1}{c}{\textbf{Precision}} & \multicolumn{1}{c}{\textbf{Recall}}  & \multicolumn{1}{c}{\textbf{F1}} \\ 
        \midrule
        \multirow{2}{*}{{\parbox{1.5cm}{\centering Dev}}}
        & FuzzyMatch & $69.3$ & $62.5$ & $65.7$\\
        & \ourmodel & $71.3$ & $70.4$ & $70.8$ \\
        \midrule
        \multirow{2}{*}{{\parbox{1.5cm}{\centering Finance}}}
        & FuzzyMatch & $35.3$ & $31.5$ & $33.2$\\
        & \ourmodel & $42.9$ & $44.7$ & $43.8$ \\
        \bottomrule
    \end{tabular}
    \caption{Results of \ourmodel grounding model compared with fuzzy string match on \fsp dev and finance splits.}
    \label{tab:ground_reults}
\end{table}

\subsection{Parser Model}\label{sec:parser}
The goal of the parser is to \ti{predict the executable SQL according to question and database schema}. The main challenge is how to model the database structure to infer the implicit schema mentioned, and how to make use of the grounded knowledge (\ti{i.e.}, knowledge-fusion) to leverage grounded knowledge.
We are inspired by the recent progress in adopting the large pre-trained language model in semantic parsing problems. 
For instance, \cite{Scholak2021:PICARD,shin-constrained,unisar,UnifiedSKG} achieves excellent performance on several semantic parsing tasks under the simple pretrained language model framework, such as BART~\cite{lewis-etal-2020-bart} and T5~\cite{2020t5}.

Given that, we propose to adopt UniSAr~\cite{unisar} as the base parser in this work.
It improves the vanilla BART with three non-invasive extensions and achieves SOTA or competitive performance on seven text-to-SQL benchmarks.
Concretely, the input of the model is the concatenation of the question, serialized schema, and retrieved formulaic knowledge.
We propose that the parser should correctly adopt the grounded formulaic knowledge during SQL generation.

\section{Experimental Results and Analysis}
To evaluate our approach: \ourmodel with external \knowledge bank, we conduct several experiments on \fsp benchmark.
We report both the overall results of the pipeline and the fine-grained results of each module.
We also conduct error analysis and categorize the bad cases into three main classes.
Note that we report the average experimental results of each setting during three runs\footnote{Code and data are available at \href{https://github.com/microsoft/ContextualSP/tree/master/knowledge_intensive_text_to_sql}{link}.}.

\subsection{Experimental Setup}
The retriever returns the top-3 retrieved formulaic knowledge items from the bank.
The grounding model further aligns the concept in formulaic knowledge into schema elements and the decision threshold $H$ is $0.6$ which is decided empirically.
The parser receives the grounded knowledge, table schema, and user query as the input and then outputs the SQL.
For the parsing baseline, we adopt UniSAr~\cite{unisar} as the vanilla parser\footnote{More implementation details could be found in Appendix~\ref{app:detail_model}}.

\subsection{Overall Results}
As shown in Tab.~\ref{tab:main_reults}, we could observe that: 
(1) \ourmodel exceeds the vanilla model by $28.2$\%, which indicates the effectiveness of using \knowledge;
(2) grounding the formulaic knowledge improves the \ourmodel by $9.0$\%;
(3) the oracle formulaic knowledge (retrieve correctly and grounding correctly) reaches the upper bound of \ourmodel $74.8$\%, which implies the potential improvement room for \fsp.

\subsection{Fine-grained Results}
We compare the retriever and grounding model with other baselines, on both the dev set and the test set of \fsp in the finance domain, to examine the performance in general and domain-specific scenarios.

\paragraph{Retriever}
We compare the retriever of \ourmodel (bi-encoder) with BM-25~\cite{bm25}.
The evaluation metric is the recall score over retrieved results.
We observe that the finance domain is more challenging than the general domain (dev split) since it contains many homogeneous formulaic knowledge items that express the same intention in the left part but with different computation ways in the right part. For example, there are two ways to compute the `EBIT' in Fig.~\ref{fig:framework}.

\paragraph{Grounding}
We compare the grounding model of \ourmodel with the fuzzy string match-based method
\footnote{It enumerates all n-gram ($n \leq 5$) of the concepts in formulaic knowledge, and links each of them to schema element by fuzzy string matching.}. 
Following the previous work~\cite{lei-etal-2020-examining,liu-etal-2021-awakening}, we report the micro-average precision, recall, and F1-score.
We could observe that:
(1) the model-based grounding improves the performance by $5.1$\% and $10.6$\% respectively;
(2) the domain-specific data like Finance poses more challenging cases than the general domain, where finance is behind the dev by about $27.0$\%.

\begin{figure*}[tb]
    \centering
    \includegraphics[width=1\linewidth]{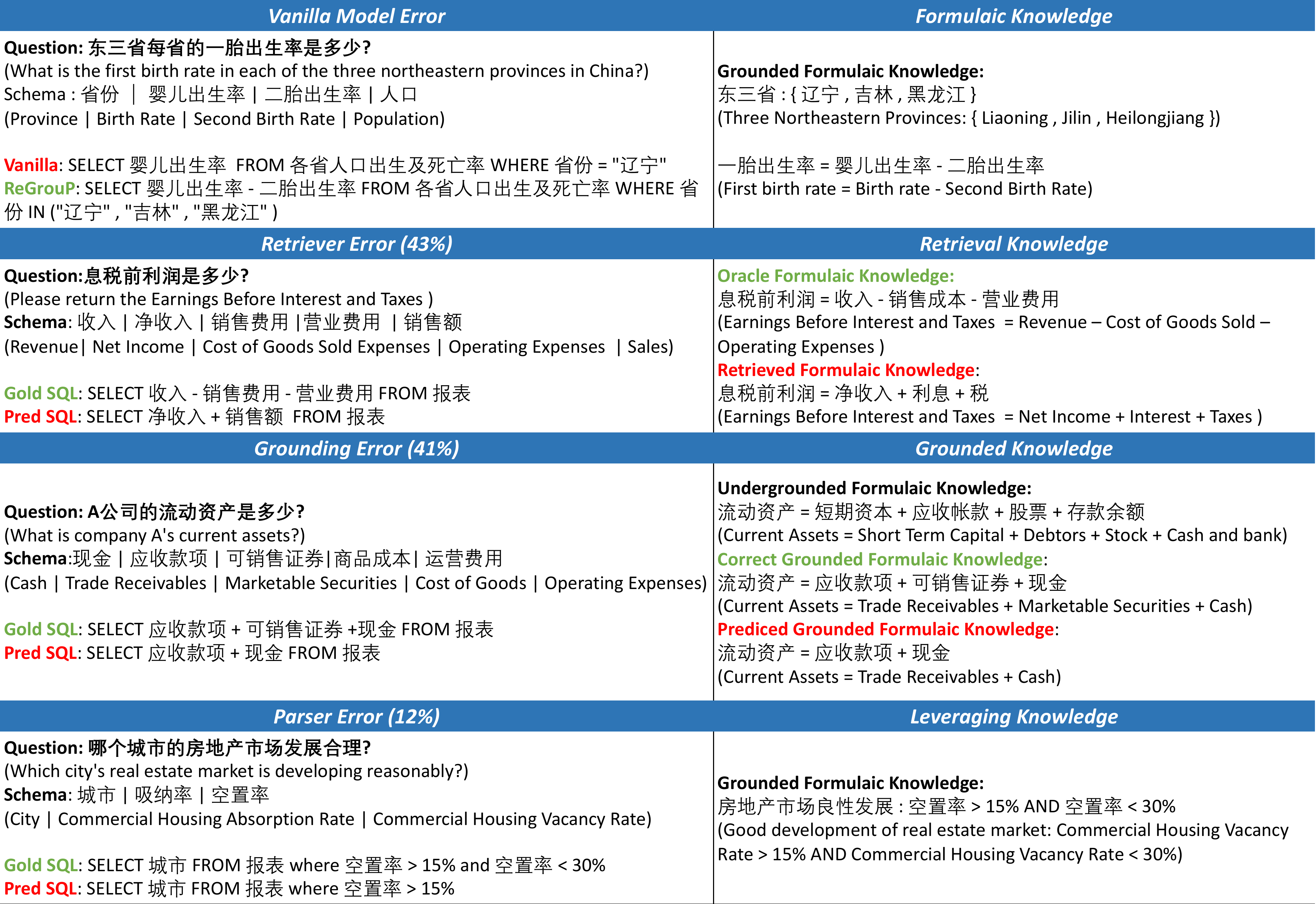}
    \caption{
        Case studies of \ourmodel. 
        We first compare it with the vanilla parsing model.
        Then we classify the bad cases of \ourmodel into three categories:
        (1) Retriever Error: \ti{not getting the knowledge from bank}; 
        (2) Grounding Error: \ti{not learning the knowledge by alignment}; 
        (3) Parser Error: \ti{not using the grounded knowledge in generation}.
    }
    \label{fig:case_study}
\end{figure*}

\subsection{Error Analysis}\label{sec:error_analysis}
We sample 300 cases from the dev split and 100 cases from finance/estate/transportation in the test split respectively (600 in total) for error analysis.

\paragraph{Vanilla Model Error} We first compare the correct case of \ourmodel while predicted incorrectly by the vanilla model. 
As the example in the first part of Fig.~\ref{fig:case_study}, the vanilla model is unable to predict the unseen operation during training. 
In contrast, the grounded \knowledge enables \ourmodel to predict the operation over schema elements correctly.

Then we categorize the error of \ourmodel into three main classes and list their percentage in Fig.~\ref{fig:case_study}. Finally, we discuss the potential future work in improving each part of \ourmodel.
An advantage of \ourmodel is the decoupled framework could track each type of bad case individually, avoiding the catastrophic forgetting problem.
\paragraph{Retrieval Error} 
About 43\% errors are attributed to the retriever where the model \ti{doesn't get the correct knowledge from bank} since it can't distinguish the semantic difference between the closed formulaic knowledge items. 
Future work should improve its distinguishing ability by fine-grained modeling, like attention mechanism~\citet{flowqa}.

\paragraph{Grounding Error}
About 41\% errors are caused by incorrect grounded knowledge where the model \ti{doesn't learn the knowledge by alignment} since it can't correctly align the concept to schema elements.
Future work should focus on how to derive the grounding information under weak supervision or even without supervision. It would greatly alleviate the severe annotation effort in specific domains.

\paragraph{Parsing Error}
There are still 8\% error cases caused by parsing, where the formulaic knowledge is correctly retrieved and grounded but the parser still \ti{doesn't use the grounded knowledge in generation} well. 
Future work should improve it by explicitly modeling the copy process of knowledge from the input to the SQL snippet position, such as implementing the additional gate mechanism.

\paragraph{Other Error} The remaining 8\% errors are about the SQL generation, such as the GROUP-BY or nested SQL.
Since it's not our main focus, we ignore these cases in Fig.~\ref{fig:case_study} for brevity.

\section{Discussion}\label{sec:discussion}
\paragraph{Is formulaic knowledge better than textual knowledge for text-to-SQL?}
In Sec.\ref{sec:motivation}, we argue that \knowledge is preferred over textual knowledge intuitively. 
Empirically, we conduct the experiments by the following steps:
(1) transforming the formulaic knowledge to textual knowledge through annotators;
(2) training the retriever and parser with textual knowledge under the same experiment setting as formulaic knowledge.
Experimental results reveal that textual knowledge receives an overall performance degradation of $13.6$\% compared with Table~\ref{tab:main_reults}. 
We conclude that \ourmodel prefers \knowledge since it's more close to the SQL snippets or schema representation.
Moreover, formulaic knowledge is both precise and concise.
In contrast, textual knowledge is redundant and much more diverse in expressing the equivalent meaning. 

\paragraph{What's the cost of collecting \knowledge?}
During the collection process of \knowledge bank (19 domains), 
we found most domains have the public knowledge resource.
Moreover, the effort spent on collection \knowledge is also acceptable compared with annotating data examples.
For example, we spent \tif{4 hours} in collecting \tif{219} formulaic knowledge in the finance domain, which is far more effective than annotating the equivalent data examples. 
Eventually, \knowledge improves the performance by $35.0$\% without retraining the model as shown in Table~\ref{tab:main_reults} (from $8.7$\% to $43.7$\%).

\paragraph{How to expand the scope of \knowledge further?}
In this paper, we mainly focus on domain knowledge and mathematical knowledge and transfer them into \knowledge format for model learning.
Other types of knowledge would improve the \kit further, such as the commonsense knowledge (\ti{e.g.}, water freezing point: temperature=0) or personalized information (\ti{e.g.}, \ti{favourite food: Tiramisu}).
Thus, we could package these types of knowledge into a formulaic format in future work.

\section{Related Work}
\subsection{Domain Generalization of Text-to-SQL}
To be applicable in real scenarios, a text-to-SQL model should generalize to new domains without relying on expensive domain-specific labeled data.
Previous work has shown that current text-to-SQL usually fails on domain generalization scenarios~\cite{finegan-dollak-etal-2018-improving}.
Recent approaches track this problem including data synthesis~\cite{yin-syn}, meta-learning~\cite{wang-etal-2021-meta} and encoder pretraining~\cite{yin-etal-2020-tabert,herzig-etal-2020-tapas}.
Most recently, ~\citet{zhao-etal-2022-bridging} proposed to adopt schema expansion and scheme pruning to preprocess the table schemas.

We highlight that compared with the schema-expansion approach, the advantage of our approach (\ourmodel with \knowledge) is the broad knowledge scope: we not only consider the calculation knowledge but also union knowledge and condition knowledge. 
Moreover, our approach is extensible with an external and maintainable \knowledge bank.

\subsection{Retrieval Enhanced Semantic Parsing}
There has been a recent trend toward leveraging retrieval-enhanced methods in various NLP tasks such as machine translation~\cite{cai-etal-2019-retrieval} and question answering~\cite{karpukhin-etal-2020-dense}.
Similar with \ourmodel, previous work \cite{gupta-etal-2022-retronlu,pasupat-etal-2021-casper} leverage a retrieval step to provides examples as the context of input for seq2seq model learning.

However, our approach differs in two ways:
(1) our retrieval object is grounded formulaic knowledge which contains more condensed information than data example;
(2) prior work directly leverage the retrieved results. We leverage the grounding model to edit the retrieved formulaic knowledge to make it more relevant to the question and schema.

\section{Conclusion and Future Work}
This paper explores \knowledge to address the \kit problem, which would advance the professional application of text-to-SQL such as data analysis for domain experts.
First, we analyze the challenge of \kit and construct a new challenging benchmark \fsp.
Then we propose to address this problem from the view of \knowledge.
Concretely, we propose a simple framework \ourmodel to leverage an external \knowledge bank.
Experimental results reveal that \ourmodel with \knowledge achieves the $28.2$\% improvements overall.

We further discuss three directions in improving the \ourmodel via analyzing different types of bad cases:
(1) iterative filling in the blank of formulaic knowledge bank; 
(2) mitigating the gap between formulaic knowledge and specific schema via improving the grounding model; 
(3) driving the parser to fully make use of more complicated (\ti{e.g.}, commonsense) formulaic knowledge.

\section*{Ethical Considerations}\label{sec:ethical_considerations}
This work presents \fsp, a free and open dataset for the research community to study the \kit problem.
Data in \fsp are constructed based on DuSQL~\cite{wang-etal-2020-dusql}
, a free and open cross-database Chinese text-to-SQL dataset.
We also collect formulaic and table data from CNKI\footnote{\href{https://oversea.cnki.net/index/}{https://oversea.cnki.net/index/}} and Baidu Wenku\footnote{\href{https://wenku.baidu.com/}{https://wenku.baidu.com/}}, which are also free and open for academic usage.
The content of the table is anonymized to address the privacy issue.
To annotate the \fsp, we recruit 3 Chinese college students (1 female and 2 males). Each student is paid 4 yuan (\$0.6 USD) for annotating the (SQL, question) pairs and 2 yuan (\$0.3USD) for collecting the formulaic knowledge items.
This compensation is determined according to the prior similar dataset construction~\cite{guo-etal-2021-chase}.
Since all question sequences are collected against open-access databases or public tables, there is no privacy issue.

\section*{Limitations}
(1) \fsp is built based on DuSQL, a Chinese large-scale text-to-SQL dataset. Thus the language coverage of this paper is limited to Chinese. We leave the extension to other languages for future work.
(2) For the scope of \knowledge, we mainly address three types of knowledge to associate with each SQL phrase: calculation, union, and condition. 
Some types of knowledge are under-explored such as commonsense knowledge.
(3) For the model design of \ourmodel, we build it from improving many existing works. Despite achieving promising evaluation results, the case studies reveal that many challenging remains during the retrieval, grounding, or parsing. 

\section*{Acknowledgement}
We thank all anonymous reviewers for their constructive comments. 
Wanxiang Che was supported via the grant 2020AAA0106501 and NSFC grants 62236004 and 61976072.
Dechen Zhan is the corresponding author.

\bibliography{anthology}
\bibliographystyle{acl_natbib}

\clearpage

\appendix

\section{Details of Formulaic Knowledge Bank}

\subsection{Knowledge Source}

We construct the formulaic knowledge bank across 19 domains following \fsp and 1 misc domain.
The misc domain stores the infrequent or general knowledge items in \fsp, such as the calculation of density, and speed.
\label{app:detail_data}
In the following, we will briefly analyze the collected bank.

\subsection{Statistic Across Domain}
\begin{figure*}[htb]
    \centering
    \includegraphics[width=1\linewidth]{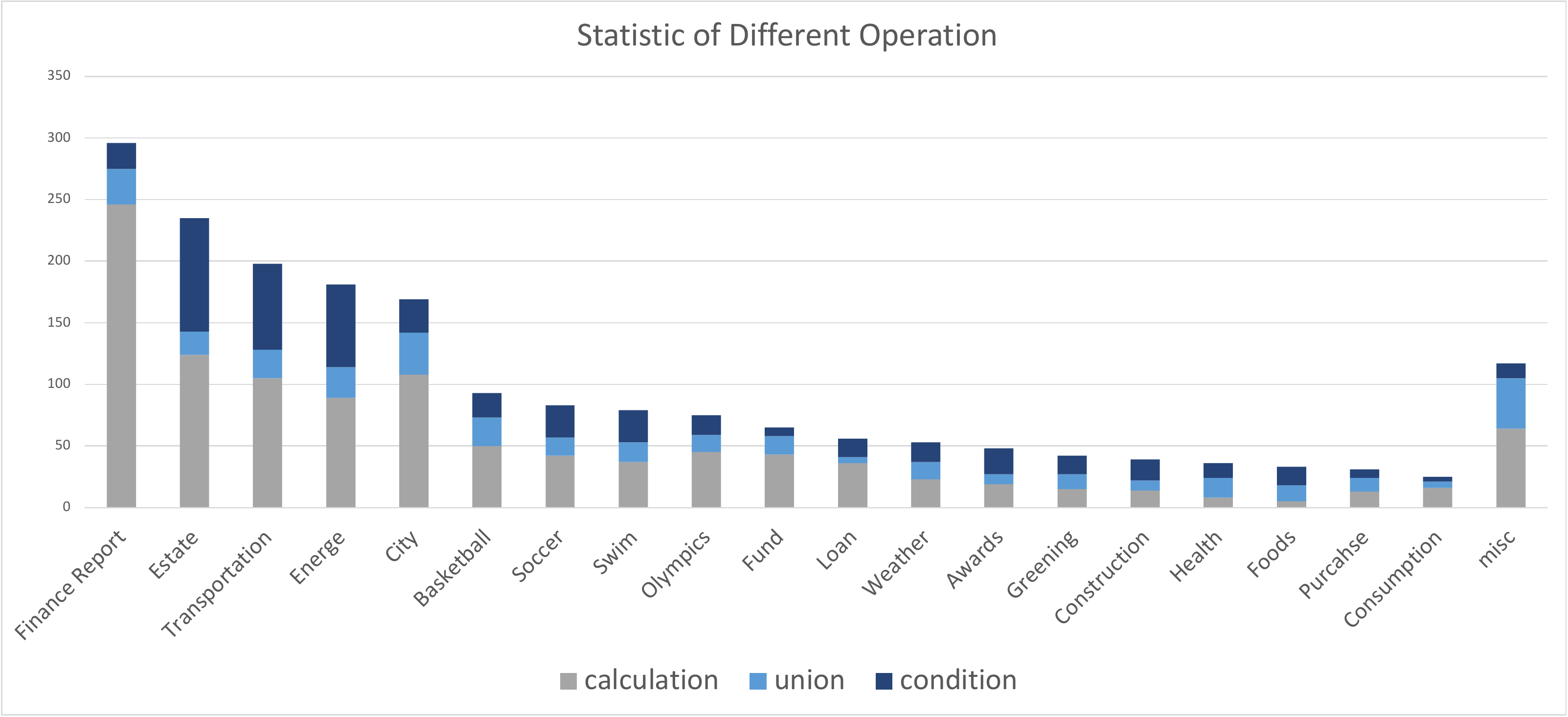}
    \caption{
        Statistic of three operations in different domains.
    }
    \label{fig:statistic_of_operation}
\end{figure*}

Different domains have different amounts of publicly available data online.
As shown in Fig.~\ref{fig:statistic_of_operation}, not unsurprisingly, finance and estate share the most plentiful publicly available resource.
% 	We attribute it's because the distribution of 

\subsection{Distribution within Domain}
We also observe the different distribution of knowledge across domains.
If the domain focus on \ti{calculation} (\ti{e.g.}, finance report and fund), we assume the data analysis tends to be more \tif{objective}, which is easier for model learning.
If the domain focus on \ti{condition} (\ti{e.g.}, estate and awards), we assume the data analysis tends to be more \tif{subjective} since it's more challenging in learning semantics.

\begin{figure*}[htb]
    \centering
    \includegraphics[width=1\linewidth]{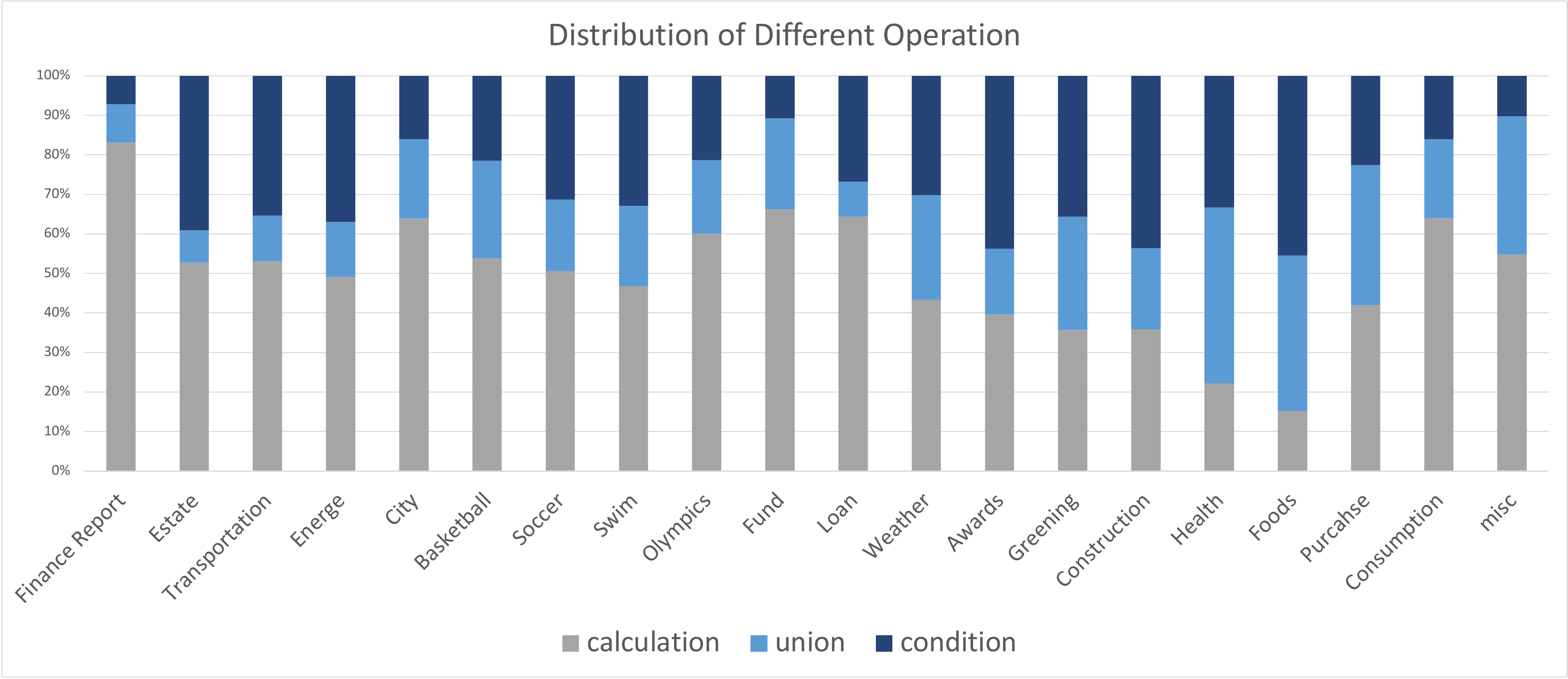}
    \caption{
        Distribution of three operations in different domains.
    }
    \label{fig:distribution_of_operation}
\end{figure*}

\section{Implementation Details of \ourmodel} \label{app:detail_model}
\paragraph{Retriever}

We implement the retriever based on the code of \citet{karpukhin-etal-2020-dense}\footnote{\href{https://github.com/facebookresearch/DPR}{Code of DPR Retrieval Model}}. 
We adopt the Chinese BERT-wwm-ext~\cite{chinese-bert} as pretrained encoder\footnote{\href{https://github.com/ymcui/Chinese-BERT-wwm}{Chinese-BERT-wwm Model}}. It would return the top-3 retrieved formulaic knowledge.
Future work could improve the negative sampling by in-batch sampling or BM25-based sampling following \citet{karpukhin-etal-2020-dense}.

\paragraph{Grounding Model}
The code of ETA\footnote{\href{https://github.com/microsoft/ContextualSP/tree/master/awakening_latent_grounding}{Code of ETA Grounding Model}} is not released at the time of submission of this paper. 
We re-implement the ETA model based on the paper using pytorch~\cite{pytorch}.
We evaluate our implemented model with the original model on \textsc{Spider-L}~\cite{lei-etal-2020-examining} to examine whether the re-implemented model works. 
Our model achieves $82.1$\% column F1 score where \citet{liu-etal-2021-awakening} reported $82.5$\%.
The experiments on \fsp also employ the Chinese BERT.

\paragraph{Parser}
We build the paper based on the code of \citet{unisar}\footnote{\href{https://github.com/microsoft/ContextualSP/tree/master/unified_parser_text_to_sql}{Code of UniSAr Parser}}
We choose the mBART-CC25\footnote{\href{https://github.com/facebookresearch/fairseq/blob/main/examples/mbart/README.md}{mBART Model}} as the base model to fine-tune.
Following the vanilla model, we build the input of parser as follows: \ti{[schema] | [grounded formulaic knowledge] | [question]}, where \ti{`|'} is the delimiter across different parts.

\paragraph{Resource and Tools}
For tokenization, we employ Stanza~\cite{qi2020stanza} considering its excellent performance.
For the retriever and grounding model, we import the BERT model with Transformer library~\cite{wolf-etal-2020-transformers}.
For parser mode, we preprocess the data and fine-tune the mBART with fairseq framework~\cite{ott-etal-2019-fairseq}

\paragraph{Device and Training Time}
We conduct all these experiments on one NVIDIA TESLA V100-32GB GPU. The training of the retriever, grounding model, and parser takes about 4 hours, 3 hours, and 8 hours respectively.
The minimum device requirement is NVIDIA TESLA P100-16G to fine-tune mBART.

\paragraph{Hyper-parameters}
All the hyper-parameters are kept the same as cited paper of each model. 
The only difference is the \ti{batch size} of the retriever and grounding model, we turn it into the maximum number to fit in the NVIDIA TESLA V100-32G GPU.

\end{document}